\documentclass{article} 
\usepackage{nips15submit_e,times,graphicx}
\usepackage{hyperref}
\usepackage{url}

\title{An Ensemble Classifier for Predicting the Onset of Type II Diabetes}

\author{ John Semerdjian \\
School of Information \\
University of California Berkeley\\
Berkeley, CA 94720 \\
\texttt{jsemer@berkeley.edu} \\
\And
Spencer Frank \\
Department of Mechanical Engineering\\
University of California Berkeley\\
Berkeley, CA 94720 \\
\texttt{spencerfrank@berkeley.edu} \\
}

\nipsfinalcopy 

\begin{document}

\maketitle

\begin{abstract}
\textit{}\href{https://youtu.be/Cq1n0dqBisc}{Short Video Abstract}  \

Prediction of disease onset from patient survey and lifestyle data is quickly becoming an important tool for diagnosing a disease before it progresses. In this study, data from the National Health and Nutrition Examination Survey (NHANES) questionnaire is used to predict the onset of type II diabetes. An ensemble model using the output of five classification algorithms was developed to predict the onset on diabetes based on 16 features. The ensemble model had an AUC of 0.834 indicating high performance. 

\end{abstract}

\section{Introduction}
Machine learning (ML) has proven itself a very useful tool in the ever expanding field of bioinformatics. It has been effectively used in the early prediction of diseases such as cancer \cite{capriotti_predicting_2006}, and work is being done in predicting the onset of Alzheimer's and Parkinson's disease \cite{capriotti_predicting_2006}. These predictions are based on data from gene sequencing and biomarkers, among other types of biological measurements. Efforts have also been made to predict the onset of diseases such as type II diabetes based on survey data. Thanks to technology that makes it much easier to collect survey data, in the future more data may become available on a larger sample of the population. This larger volume of survey data presents a new opportunity to improve overall prediction of disease, especially in diseases where lifestyle is  highly correlated to disease onset \cite{_reduction_2002}.

Previous work has focused on the use of survey data to predict the onset of diabetes in a large sample of the population using Support Vector Machines (SVM) \cite{yu_application_2010}, obtaining an area under the receiver operating characteristic (ROC) curve of 0.83. The dataset used in \cite{yu_application_2010} is called the National Health and Nutrition Examination Survey (NHANES) \cite{center_for_disease_and_control_national_????}. Our goal is to use this same dataset and attempt to improve prediction accuracy. A secondary objective is to more clearly interpret the results, and indicate how such a model could used in the real world to improve preventative care.

This paper is organized as follows. First, we will describe the NHANES dataset to be analyzed in this study. Next a description of the implemented classification methods will be described, along with a description of the proposed ensemble model. Results of each proposed model will be presented that indicate performance. Model performance will be discussed as well as some real world applications of the proposed classification models. 

\section{NHANES Survey Data, Feature Section, and Label Determination}
The National Health and Nutrition Examination Survey (NHANES) data is an on-going cross sectional sample survey of the US population where information about participants is gathered during in-home interviews. In addition to the in-home surveys, participants also have the option to obtain a physical examination in mobile examination centers \cite{center_for_disease_and_control_national_????}.

\subsection{Patient Exclusion and Label Assignment}
As in \cite{yu_application_2010}, we limited our study to non-pregnant participants over 20 years of age. We focused our efforts on three waves of the survey conducted between 1999-2004, given the consistency across survey questions between waves from this time period.

Our problem is a binary classification, and as such we must assign labels to our samples. To do this we used measurements from each sample. The first was the patient's answer to the question, ``Has a doctor ever told you that you have diabetes?" If that patient answered `yes,' than a `1' was assigned to the patient indicating that they had type II diabetes, and was `0' otherwise. Using patients that responded to this question produced about 900 samples. The second, and more common method for determining the label of a patient was the value of their plasma glucose level, tested during examination. If the patient's glucose level was greater than 126 mg/dL, they were labeled as diabetic and given a `1'. If their glucose level was less than this threshold they were assigned to the non-diabetic group with a `0'. This method of labeling produced the remaining 4600 samples. With the labels determined, we now move onto feature selection. 

\subsection{Feature Selection}
The features were chosen to be similar to the features used in \cite{yu_application_2010}, since much effort was already put in during this previous study to select good features. Additional features were examined, such as diet, however upon further analysis most of these were excluded due to their negligible effect on performance or a significant portion of feature data was missing ($>$ 60\% missing). Two features that were added to the data were cholesterol and leg length, which were found to be important features in \cite{heredia2013genetic}. Of the multitude of potential features, we chose to focus our attention on only 16 features. These 16 features are described in Table \ref{feature-table}. This made interpretation simpler and training-time shorter. The training-time was a major concern because of the many parameters to be optimized during cross-validation.

\subsection{Cross-Validation and Test Set}
After exclusion and labeling, 5515 total samples were available from the NHANES 1999 to 2004 dataset. Before model training, a 20\% test set was removed from the entire dataset. This left 4412 samples for training and 1103 for testing. 

For hyperparameter tuning 10-fold cross validation was used, with grid-search. To obtain testing accuracy error, bootstrapping was utilized. 

\begin{table}[h!]
\caption{Feature Description Table with Feature Importance Ranked Based on Random Forest Classifier}
\begin{center}
\begin{tabular}{llll}
\multicolumn{1}{c}{\bf Feature Name}  &\multicolumn{1}{c}{\bf DESCRIPTION} &\multicolumn{1}{c}{\bf NHANES Code} &\multicolumn{1}{c}{\bf Feature Importance}
\\ \hline \\
AGE        &Age at time of screening & RIDAGEYR & 0.169\\
WAIST      &Waist circumference (cm) & BMXWAIST & 0.107\\
REL        &Blood relatives have diabetes? & MCQ250A & 0.087\\
HEIGHT     &Height (cm) & BMXHT & 0.086\\
CHOL       &Total cholesterol (mg/dL) & LBXTC & 0.085\\
LEG        &Upper Leg Length (cm) & BMXLEG & 0.080\\
WEIGHT     &Weight (kg) & BMXWT & 0.071\\
BMI        &Body mass index (kg/m$^2$) & BMXBMI & 0.067\\
RACE       &Race/ethnicity & RIDRETH1 & 0.050\\
HBP        &High blood pressure? & BPQ020 & 0.046\\
INCOME     &Annual household income (dollars) & INDHHINC & 0.039\\
ALC        &Amount of alcohol in past year? & ALQ120Q & 0.038\\
SMOKE      &Age started smoking cigarettes regularly & SMD030 & 0.036\\
EDU        &Education level & DMDEDUC2 & 0.017\\
EXER       &Daily physical activity level & PAQ180 & 0.012\\
GEND       &Gender & RIAGENDR & 0.011
\end{tabular}
\end{center}
\label{feature-table}
\end{table}

\section{Methods}

\begin{figure}

\begin{center}
\framebox[5.5in]{\includegraphics[scale=1]{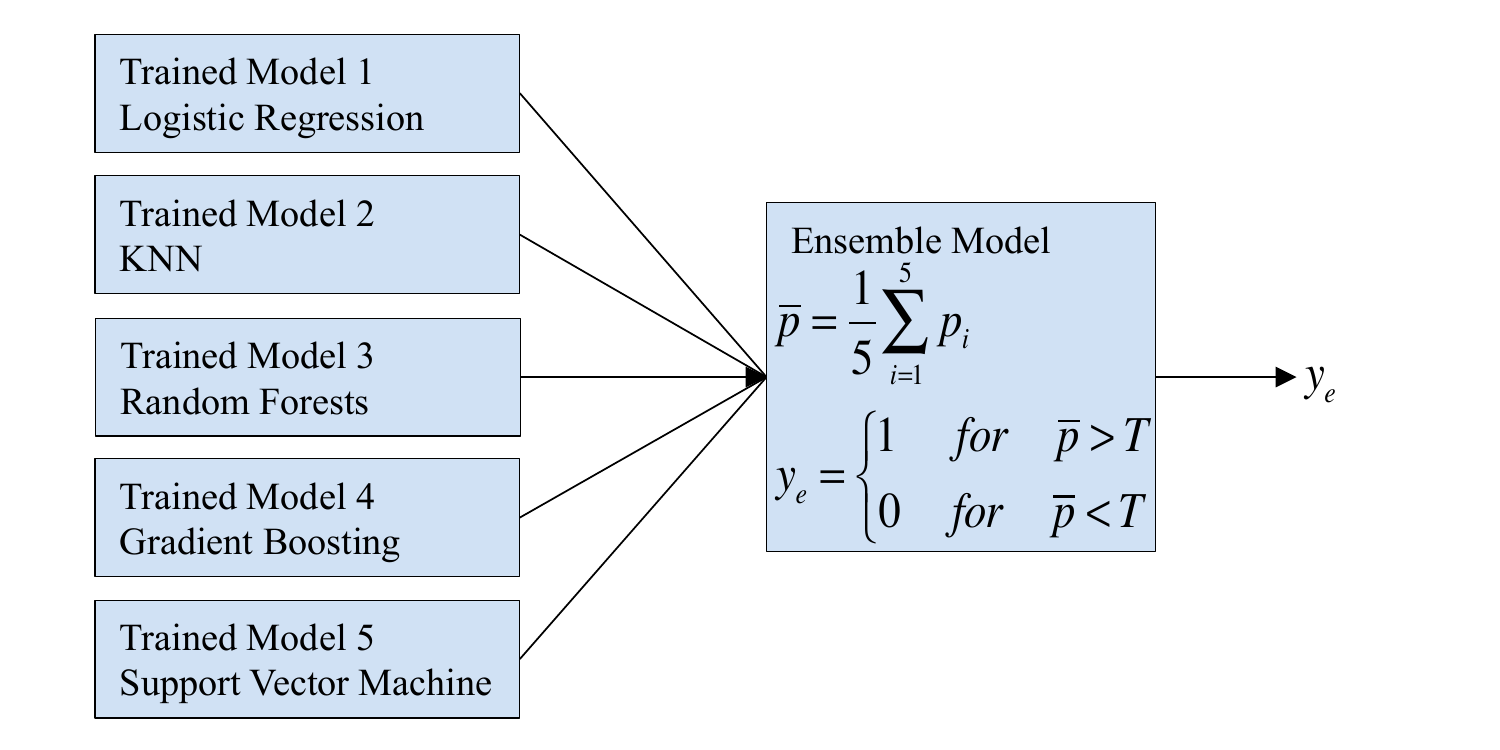}}
\end{center}
\caption{Schematic of Ensemble Model Classifier. Each trained model outputs a probability $p$. The ensemble model takes each of these probabilities and calculates the unweighted average probability, $\bar{p}$. Finally, the decision boundary T is selected based on tuning from recall results. Initially T=0.5, but by adjusting T, higher diabetes prediction rates can be obtained at the expense of labeling more healthy patients as diabetics.}
\label{ensemble_scematic}
\end{figure}

All modeling was done using scikit-learn, a python-based toolkit for simple and efficient machine learning, data mining and data analysis \cite{scikit-learn}. 

\subsection{Individual Models}
Five simple models were chosen that presented fairly high discriminative power on their own: Logistic Regression, K-Nearest Neighbors (KNN), Random Forests, Gradient Boosting, and Support Vector Machine (SVM). Other models such as Naive Bayes were tried, but did not produce accurate results compared to the five previous models. 

Each model had a few hyperparameters associated with it. The five models in total contained a total of 10 hyperparameters. Using grid search would be intractable to tune all the hyperparameters simultaneously, since to test even just 10 values for each would produce 10$^{10}$ evaluations. Instead, each model was tuned separately, with the result being shorter tuning times that were possible on a laptop computer. 

\subsection{Ensemble Model}
After training, the five previously described models were set to output probabilities $p$. These probabilities were then fed into the ensemble model (see Figure \ref{ensemble_scematic}), which took an unweighted average of the inputs to obtain $\bar{p}$. Initially, as is typical, T was set to 0.5. This however lead to a very poor recall rate on the diabetic patients. Because of this observation of very high type II error (i.e. low recall), and considering that it is of greater importance to identify a diabetic than misidentify a healthy patient, the decision boundary T was adjusted to obtain a more desirable recall rate.

\section{Results}

\subsection{Overall Discriminative Power of All Models}
\begin{figure}[t]
\begin{center}
\framebox[4.5in]{\includegraphics[scale=0.5]{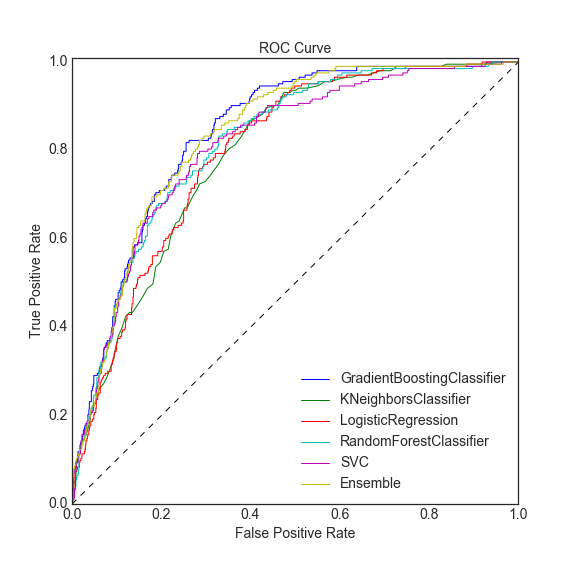}}
\end{center}
\caption{Receiver Operating Characteristic (ROC) curves are plotted for each individual model and the ensemble model. The dotted line represents random-guessing. The performance metrics can be found in Table \ref{metrics_table}.}
\label{roc_individual}
\end{figure}

The overall discriminative power of the model is shown by the Receiver Operating Characteristic (ROC) curves in Figure \ref{roc_individual}. From the figure it can be seen that the Gradient Boosting Classifier performs best with an AUC of 0.84. Random forests also performs exceptionally well. The other three classifiers were noticeably worse at classification, with the worst performance being the KNN. The AUCs of each individual model, as well as other metrics, can be found in Table \ref{metrics_table}. 

Unfortunately, the ensemble model failed to significantly improve performance. In fact it was less accurate than the Gradient Boost Classifier as measured by AUC. However the difference between the two was very small, as seen in the performance metrics in Table \ref{metrics_table}. 

\begin{table}[h]
\caption{Performance Metrics for the Five Models, and the Ensemble Model}
\begin{center}
\begin{tabular}{lllll}
\multicolumn{1}{c}{\bf Model} &\multicolumn{1}{c}{\bf AUC} &\multicolumn{1}{c}{\bf F1 Score} &\multicolumn{1}{c}{\bf Recall} &\multicolumn{1}{c}{\bf Precision}
\\ \hline \\
K-Nearest Neighbors & 0.793 & 0.73 & 0.82 & 0.67 \\
Logistic Regression & 0.797 & 0.79 & 0.81 & 0.78 \\
Support Vector      & 0.812 & 0.79 & 0.82 & 0.78 \\
Random Forests      & 0.817 & 0.79 & 0.82 & 0.79 \\
Gradient Boosting   & 0.840 & 0.81 & 0.82 & 0.80 \\
Ensemble            & 0.834 & 0.78 & 0.82 & 0.78 \\
\end{tabular}
\end{center}
\label{metrics_table}
\end{table}

\subsection{Recall and the Decision Boundary T}
\begin{figure}[h]
\begin{center}
\framebox[5.5in]{\includegraphics[scale=0.5]{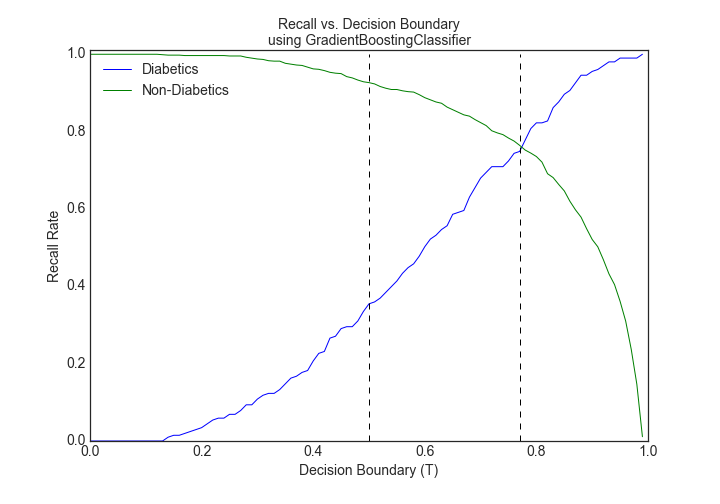}}
\end{center}
\caption{Recall vs. Decision Boundary T curves for diabetics and non-diabetics for the top performing classifier, Gradient Boosting. At the default classification of T=0.5 (middle dotted line), the recall rate is 0.93 for non-diabetics and 0.35 for diabetics. For T=0.78 (right dotted line) the recall rate for non-diabetics is 0.75 and the recall rate for diabetics is 0.75, representing a large improvement in identifying the diabetics patients with only a small sacrifice in recall rate (18\%) for the non-diabetics.}
\label{recall_decision}
\end{figure}

\begin{figure}[h]
\begin{center}
\framebox[5.5in]{\includegraphics[scale=0.5]{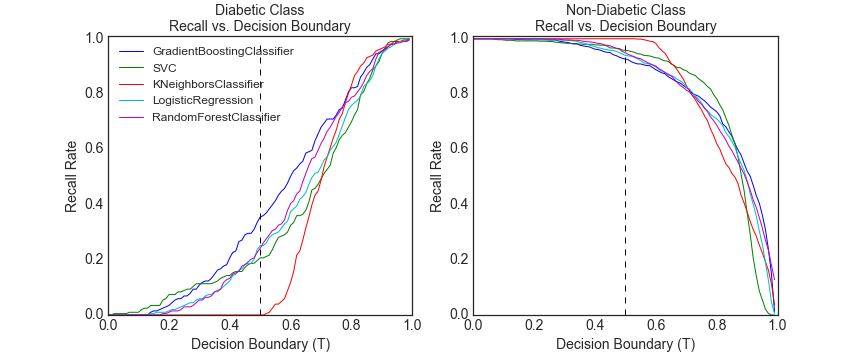}}
\end{center}
\caption{Recall vs. Decision Boundary curves for diabetic and non-diabetics by model. The Gradient Boosting Classifier had the highest performance, with the KNN classifier having the poorest performance. The other three models had similar performance. Metrics are shown in Table \ref{metrics_table}.}
\label{recall_score_all}
\end{figure}

The recall rate for the diabetics is defined as the number of diabetics identified, over the total number of diabetics in the sample. The recall rate is a function of the decision boundary T, and is plotted for the best performing classifier, Gradient Boosting, in Figure \ref{recall_decision}. Recall was also plotted for all the models for each class in Figure \ref{recall_score_all}. At the default decision boundary of T=0.5, the recall rate for diabetics was only 0.35, meaning that only 35\% of the time we classify a diabetic patient that in-fact has diabetics. Since our objective is to maximize our ability to identify diabetics, we decided to shift our decision boundary to a point which would allow us to identify diabetics at least 75\% of the time. To achieve this, we shifted out decision boundary rightward to T=0.78, where the recall rate for diabetics was 0.75, and coincidentally, the recall rate for non-diabetics was also 0.75. This means that by shifting our decision boundary, we were able to identify many more diabetics, only having to sacrifice a small decrease in recall rate of 18\% for the non-diabetics. 

\subsection{Testing Error}
\begin{figure}[h]
\begin{center}
\framebox[5.5in]{\includegraphics[scale=0.5]{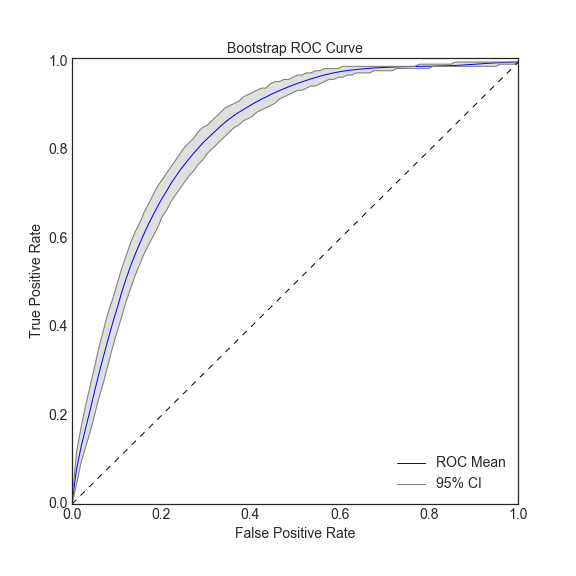}}
\end{center}
\caption{ROC curve for the Gradient Boosting Classifier. Also plotted are the 2-standard deviation spread representing the testing performance for models trained on $N_{boot}$ bootstrapped samples. The very small difference between the average (middle curve) and the top and bottom curves show that the model performance is not sensitive to the training data. The dotted line is the performance associated with random guessing. }
\label{roc_error_decision}
\end{figure}

In order to obtain the error on the test accuracy results, bootstrapping was performed for training on the best performing model, the Gradient Boosting Classifier. This produced $N_{boot}$ classifiers, all having slightly different discriminative abilities. To show the error in classification, ROC curves with continuous upper and lower error bars are shown in Figure \ref{roc_error_decision}. This figure plots three ROC curves, the middle one is the average ROC curve for all the $N_{boot}$ trained models. The upper and lower curves are the 2.5\% and 97.5\% empirical quantiles of the bootstrap sample (i.e. 95\% confidence intervals). To obtain this distribution, $N_{boot}$ = 1000 models were trained. Mean AUC from bootstrapping was 0.83, with upper and lower confidence intervals of 0.82 and 0.84. This indicates that the training is not highly sensitive to the training data.

\section{Discussion and Applications}

\subsection{Model Performance}

Five models were used to classify diabetics and non-diabetics based on survey data collected from NHANES. These five models were given equal weighting in an ensemble model that used the probabilistic output of each model. From the ROC curves of all the models (seen in Figure \ref{roc_individual}) it was found that the best performing model (the highest AUC, seen in Table \ref{metrics_table}) was the Gradient Boosting Classifier, which actually beat out the ensemble method. 

At first glance, it is surprising that the lone Gradient Boosting Classifier performed best, as it was thought that the wisdom of each model would combine into the best classifier through an ensemble method. This would have been consistent with the theory of the `wisdom of crowds.' 
The poorer performance of the ensemble method could be due to a few factors. The first is that there was inadequate hyperparameter tuning. To prevent the need to tune 10 hyperparameters simultaneously, which would have taken an exorbitant amount of time, we instead chose to tune each model separately. In doing this we prevented the ensemble model from really using the best features of each model, as we did not allow for the tuning of hyperparameters based on ensemble performance. An additional hyperparameter that should have been optimized was the weighting of each model. Our ensemble model weighted each model equally, which may have given too much weight to poorer performing models such as the KNN Classifier (the AUC for this model was much lower than the AUC of the other 4 models). 

In the future, if computational time were not an issue, simultaneous tuning of hyperparameters would be carried out. Additional performance enhancement could also be achieve by implementing a `stacked model,' which is essentially a two-layered model that would use the output of each individual model as a feature to train a lower dimensional model. 

One of our preprocessing steps was to impute values for missing data. This is particularly important because $>$25\% of cells are missing values. For numerical features (e.g. BMI and height) we calculated the mean from the training data and assigned it to missing fields in both training and test sets. For categorical variables (e.g. income bracket and alcohol use) we assigned missing values the most common label from the training set. Further performance enhancement could also be obtained by imputing missing data by building a model from the available data to predict missing values for a given feature and through matrix factorization. 

The top AUCs from the models presented in this paper were consistent with the work of \cite{yu_application_2010}, who achieved and AUC of 0.83 for a well-tuned support vector machine. Our consistency with this previous study indicates that the 0.83 AUC may be a hard upper limit on the discriminative power of models trained on the NHANES data for the purpose of the prediction of diabetes using simple classification techniques such as those discussed in the present work and in \cite{yu_application_2010}. However, considering only marginal effort was made to impute data in either works, clever data imputation may present the best path for increasing performance.

\subsection{Applications} 
When creating classification models for disease detection it is always important to keep the end use in mind. The classification model presented here is used for the purpose of early detection of type II diabetes based on easily obtainable survey data. The features are responses to questions and simple examination measurements shown in Table \ref{feature-table}. 

A decision boundary of T=0.78 was chosen that produced a recall rate of 75\% for diabetics, meaning that 75 out of 100 diabetics were correctly identified. This was considered an acceptable outcome from a diagnostic standpoint. However, the caveat is that to ensure a 75\% recall rate in diabetics, many non-diabetics will be identified as diabetics. To give specific numbers, with a recall rate of 75\% for non-diabetics, we can eliminate 5515 * 0.81*0.75 = 3350 patients from the total population of 5515 (0.81 is the proportion of non-diabetics in the entire training set).  This would leave 2165 patients left, which would be the number of patients that would need to be notified they were diabetic in order to ensure 75\% of the actual diabetics are notified. This may provide a significant decrease in over healthcare cost, by increasing preventative care for the notified patients. With this possibility in mind, a public health expert can decide whether this classification scheme is cost-effective.
  
\section{Conclusions}
Five separate models were used to classify diabetics vs. non-diabetics. The output of these models was used to create an ensemble model, which had 0.834 AUC. The best performing model was not the ensemble model, but rather the Gradient Boosting Classifier, which obtained 0.84 AUC. This AUC is consistent with previous findings and indicates that in using survey data one can expect reasonable classification performance. 

\subsubsection*{Acknowledgments}
We would like to thank Professor Alexei Efros (UC Berkeley) and Professor Isabelle Guyon (visiting professor at UC Berkeley) for for their input to this work. 

\bibliography{references.bib}
\bibliographystyle{plain}

\end{document}